\pdfoutput=1

\documentclass[11pt]{article}

\usepackage{acl} 

\usepackage{times}
\usepackage{latexsym}

\usepackage[T1]{fontenc}

\usepackage[utf8]{inputenc}

\usepackage{microtype}

\usepackage{inconsolata}

\usepackage{graphicx}
\usepackage{xcolor}
\usepackage{tikz}
\usetikzlibrary{calc}
\usepackage{colortbl}
\usepackage{booktabs}
\usepackage{todonotes}
\usepackage[most]{tcolorbox}
\tcbuselibrary{listingsutf8}
\title{Evaluating Structured Output Robustness of Small Language Models for Open Attribute-Value Extraction from Clinical Notes}

\author{
  \textbf{Nikita Neveditsin\textsuperscript{1}},
  \textbf{Pawan Lingras\textsuperscript{1}},
  \textbf{Vijay Mago\textsuperscript{2}}
\\
\textsuperscript{1}Saint Mary's University, Halifax, Canada \\
\textsuperscript{2}York University, Toronto, Canada
\\
}

\definecolor{LightGreen}{rgb}{0.88,1,0.88}
\definecolor{LightRed}{rgb}{1,0.88,0.88}

\begin{document}
\maketitle

\begin{abstract}
We present a comparative analysis of the parseability of structured outputs generated by small language models for open attribute-value extraction from clinical notes. We evaluate three widely used serialization formats: JSON, YAML, and XML, and find that JSON consistently yields the highest parseability. Structural robustness improves with targeted prompting and larger models, but declines for longer documents and certain note types. Our error analysis identifies recurring format-specific failure patterns. These findings offer practical guidance for selecting serialization formats and designing prompts when deploying language models in privacy-sensitive clinical settings.
\end{abstract}

\section{Introduction}

Structured information extracted from clinical narratives enhances clinical decision-making, streamlines reporting, and facilitates research database development \cite{wang2018clinical, garg2021role}. Small language models (SLMs) \cite{schick-schutze-2021-just} can be deployed on local hardware and therefore meet privacy requirements \cite{neveditsin2025clinical}, but their utility depends on producing outputs that downstream software can parse automatically.

This work examines \textbf{open attribute-value extraction}, a task in which an SLM identifies clinically relevant attribute-value pairs \emph{without a predefined schema} and serializes them in a standard format \cite{Banko, zheng2018opentag, li2023attgen, Brinkmann}. We compare three commonly used formats: JSON, YAML, and XML, and assess robustness via \textit{parseability}, defined as the proportion of outputs that can be successfully validated by a standard parser without manual correction. We further analyze how document length, note type, model size, and extraction scope (open vs.\ targeted for medications, symptoms, and demographics) affect parseability, and report on common structural failure modes and key interactions among these factors.

Our contributions are as follows:  (i) to the best of our knowledge, we provide the first comparative analysis of structured output parseability across three widely used serialization formats (JSON, YAML, XML) in the context of open attribute-value extraction from clinical notes;  (ii) we demonstrate how model size, prompt specificity, and clinical document characteristics systematically influence structural robustness;  (iii) we identify and categorize recurrent structural failure modes, offering practical insights into common format-specific vulnerabilities in SLM-generated outputs.

\setlength{\parskip}{0pt}




\section{Related Work}

Prior work on structured information extraction with transformer-based language models has highlighted both their semantic potential and their syntactic fragility. Research in this area can be broadly categorized by its primary evaluation focus: studies that prioritize the semantic accuracy of the extracted content, and those that more directly engage with the technical challenge of ensuring syntactic validity.

In high-stakes domains such as clinical medicine, the evaluation emphasis is typically on semantic accuracy. For example, \citet{balasubramanian2025leveraging} evaluated the extraction of 51 features from breast cancer pathology reports by comparing model outputs against expert-annotated gold standards. Similarly, \citet{kadhim2025application} measured the correctness of extracted findings in inflammatory bowel disease reports using F1 scores. In both cases, models like LLaMA-3.3 were assessed primarily on their ability to extract correct clinical content. Syntactic validity, such as whether outputs conformed to a given format, was assumed rather than explicitly evaluated. Other studies, such as \citet{elnashar2025enhancing}, explored prompt design and efficiency trade-offs across JSON, YAML, and hybrid CSV formats using GPT-4o. While they validated attribute-level correctness, structural robustness was not a primary focus.

This focus on semantics often coexists with an implicit acknowledgment of the syntactic fragility of unconstrained model outputs. Work in scientific and technical domains has more directly quantified this issue. \citet{Dagdelen2024}, in the context of materials science extraction, noted parse failures under token limits. \citet{schilling2024text} advocates constrained decoding to restrict the model's vocabulary during generation to enforce structural compliance. While this technique improves parseability, \citet{tam-etal-2024-speak} have shown that tighter constraints may also reduce reasoning flexibility, underscoring a trade-off between structural validity and expressiveness.

These findings indicate a gap in evaluating the syntactic reliability of structured outputs. Our study addresses this by focusing specifically on parseability as the primary evaluation criterion, using small instruction-tuned models.

\section{Methodology}
\subsection{Models}

To assess the impact of output format on small language models, we evaluate seven open-weight instruction-tuned models (Table~\ref{tab:models}).

\begin{table}[h!]
\centering
\renewcommand{\arraystretch}{0.9}

\resizebox{1\linewidth}{!}{ 
\begin{tabular}{lccc}
\toprule
\textbf{Model} & \textbf{Vendor} & \begin{tabular}[c]{@{}c@{}}\textbf{Params} \\ \textbf{(B)}\end{tabular} & \begin{tabular}[c]{@{}c@{}}\textbf{Ctx.} \\ \textbf{Window}\end{tabular} \\
\midrule
\texttt{Phi-4} \cite{abdin2024phi4technicalreport}                   & Microsoft     & 14   & 16K   \\
\texttt{Phi-3.5-mini} \cite{abdin2024phi3technicalreporthighly}            & Microsoft     & 3.8  & 128K  \\
\texttt{Llama-3.2-3B} \cite{grattafiori2024llama3herdmodels}         & Meta          & 3    & 128K  \\
\texttt{Llama-3.1-8B}  \cite{grattafiori2024llama3herdmodels}           & Meta          & 8    & 128K  \\
\texttt{Mistral-8B} \cite{jiang2023mistral7b} & Mistral AI    & 8    & 128K  \\
\texttt{Qwen3-4B} \cite{qwen3_2024} & Alibaba       & 4    & 32K  \\
\texttt{Qwen3-14B}  \cite{qwen3_2024}             & Alibaba       & 14   & 128K  \\
\bottomrule
\end{tabular}
}
\caption{SLMs evaluated in this study.}
\label{tab:models}
\end{table}

We selected 7 models from 4 vendors (Microsoft, Meta, Mistral, Alibaba), some of which contributed more than one model. This allowed us to reduce provider-specific bias while also covering a range of model sizes (3–14B parameters) and context window capacities (ranging from 16K to 128K tokens, as shown in Table~\ref{tab:models}). All models are openly available, support local deployment, and are widely used in the open-source community, ensuring relevance, reproducibility, and suitability for privacy-sensitive clinical use.

\subsection{Data}

We use the \textbf{EHRCon} \cite{goldberger2000physiobank, kwon2025ehrcon} dataset, a standardized, open, and ethically compliant subset of MIMIC-III \cite{Johnson2016} that supports reproducible research. It includes 105 randomly selected, de-identified clinical notes with 4,101 annotated entities mapped to 13 structured EHR tables. Derived from a large critical care database, EHRCon captures the complexity of real-world clinical documentation. Its public availability and prior ethical clearance make it suitable for secondary analysis without requiring additional ethical review. EHRCon is well-suited for evaluating structural parseability, and its detailed attribute-level annotations offer opportunities for future research on semantic validity, though we do not pursue that direction in this work.

The dataset includes three note types: discharge summaries, nursing notes, and physician notes, each with distinct content and length characteristics (Table~\ref{tab:note_stats}). Discharge summaries, the longest (avg. 1300 words, 2700 tokens), provide a comprehensive account of the hospital stay. Physician notes, of moderate length, focus on assessments and treatment plans. Nursing notes, the shortest, document vitals, patient behavior, and routine care. 


\begin{table}[h!]
\centering

\resizebox{0.9\linewidth}{!}{ 
\begin{tabular}{lrrr}
\toprule
\textbf{Type} & \textbf{\# Documents} & \textbf{Avg. Words} & \textbf{Avg. Tokens} \\
\midrule
Discharge  & 38  & 1306.47 & 2764.46 \\
Nursing    & 36  & 490.33  & 1153.63 \\
Physician  & 33  & 669.91  & 1914.93 \\
\bottomrule

\end{tabular}

}
\caption{Descriptive statistics of clinical note types.}
\label{tab:note_stats}
\end{table}

Token counts are computed by applying each model's tokenizer to every document and averaging across models from Table~\ref{tab:models}.

\subsection{Experimental Setup}

We assess SLMs in two extraction scenarios. The \textit{open} format scenario prompts the model to extract any medically relevant information it can infer from a note without relying on a predefined schema. This reflects exploratory or retrospective use cases where schema coverage may be incomplete or unavailable.
The \textit{targeted} scenario narrows the prompt to a specific category: medications, symptoms, or demographics. These categories are commonly prioritized in clinical information extraction for their central role in decision support and downstream clinical tasks \cite{sohn2013comprehensive, wang2018clinical}. This allows us to assess whether more constrained prompts yield more structurally consistent outputs.

\begin{figure}[h!]
    \centering
    \includegraphics[width=0.5\textwidth]{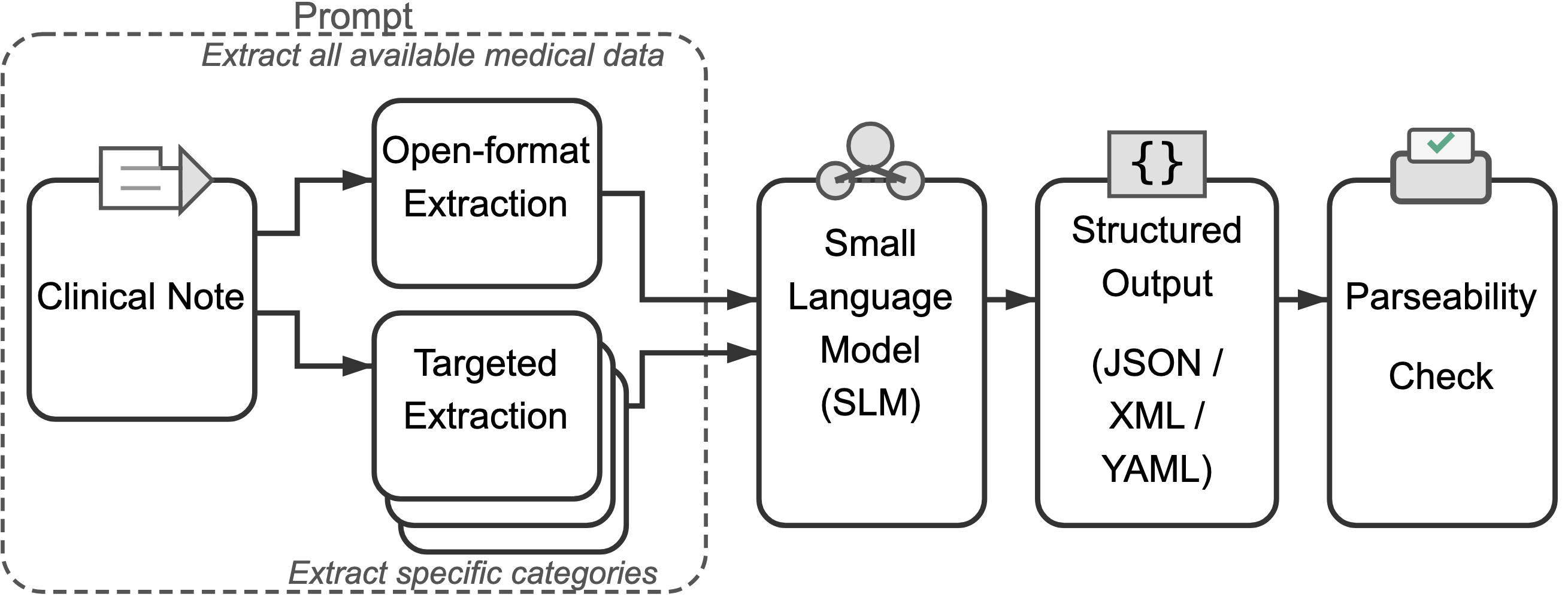}
    \caption{Workflow for evaluating structured output generation}

    \label{fig:workflow}
\end{figure}

Figure~\ref{fig:workflow} illustrates the overall workflow. A clinical note is processed under one of the two prompting conditions, passed to an SLM, and rendered in JSON, YAML, or XML. The output is then evaluated for parseability using a standard parser.

In both scenarios, we focus on parseability; we do not evaluate content accuracy. Formally, for a given model, prompt type, and a set of documents \( D \), we define the \textbf{parseability rate} as
\[
\rho(D) = \frac{n_v}{|D|},
\]
where \( n_v \) denotes the number of documents in \( D \) whose outputs were successfully parsed by a standard parser under that model and prompt type. To support our findings, we apply appropriate statistical tests. Appendix~\ref{sec:appendix1} provides additional details on the experimental setup.


\section{Results}


Table~\ref{tab:parseability} presents parseability rates across JSON, YAML, and XML for all models listed in Table~\ref{tab:models}, evaluated on the full clinical document set. Each model appears in two rows, corresponding to the open-ended and targeted extraction settings (the \textit{Setting} column). 

Parseability tends to improve with model size. To assess this effect, we grouped models by parameter count into three categories: \textit{Small} (3-4B), \textit{Medium} (8B), and \textit{Large} (14B). A Chi-squared test of independence confirmed a significant association between model size and parseability (\( \chi^2 = 106.72 \), \( p \ll 0.05 \)). Average parseability rates rose with size: Large models achieved 90.3\%, followed by Medium (82.6\%) and Small (80.9\%). The effect size, measured by Cramér’s \( V = 0.11 \), suggests a statistically significant but modest association between model size and parseability.

Prompt specificity was also a significant factor. Targeted prompts substantially boosted parseability across all formats, especially for YAML, which performs poorly in the open setting. A Chi-squared test confirmed a strong association between prompt type and parseability (\( \chi^2 = 1579.41 \), \( p \ll 0.05 \)). Cramér’s \( V = 0.42 \) indicates a medium-to-large impact of prompt type on structural validity.

\begin{table}[h!]
\centering
\renewcommand{\arraystretch}{1.0}

\resizebox{1\linewidth}{!}{%
\begin{tabular}{llccc}
\toprule
\textbf{Model} & \textbf{Setting} & \textbf{JSON} & \textbf{XML} & \textbf{YAML} \\
\midrule
\rowcolor{gray!10}
Llama-3.1-8B & Open & \textbf{59.8} & 54.2 & \textit{23.4} \\
\rowcolor{gray!10}
Llama-3.1-8B & Targeted & \textbf{97.8} & 96.9 & \textit{92.2} \\
Llama-3.2-3B & Open & \textbf{73.8} & 41.1 & \textit{29.9} \\
Llama-3.2-3B & Targeted & \textbf{94.4} & 81.6 & \textit{75.1} \\
\rowcolor{gray!10}
Mistral-8B & Open & \textbf{81.3} & 57.9 & \textit{47.7} \\
\rowcolor{gray!10}
Mistral-8B & Targeted & \textbf{96.0} & 89.1 & \textit{80.4} \\
Phi-3.5-mini & Open & \textbf{83.2} & \textit{43.0} & 52.3 \\
Phi-3.5-mini & Targeted & \textbf{99.4} & 94.7 & \textit{83.5} \\
\rowcolor{gray!10}
Phi-4 & Open & \textbf{100.0} & 61.7 & \textit{44.9} \\
\rowcolor{gray!10}
Phi-4 & Targeted & \textbf{100.0} & 98.4 & \textit{97.8} \\
Qwen3-14B & Open & \textbf{98.1} & \textit{43.0} & 47.7 \\
Qwen3-14B & Targeted & \textbf{99.4} & \textit{97.2} & 97.5 \\
\rowcolor{gray!10}
Qwen3-4B & Open & \textbf{95.3} & 39.3 & \textit{29.0} \\
\rowcolor{gray!10}
Qwen3-4B & Targeted & \textbf{97.2} & 94.4 & \textit{86.3} \\
\bottomrule
\end{tabular}
}
\caption{Parseability rates (\%) by model and output format across the full document set. Each model appears in two rows, corresponding to open-ended and targeted extraction settings (prompt types). \textbf{Bold} indicates the highest parseability per row; \textit{italic} indicates the lowest.}

\label{tab:parseability}
\end{table}

To test for the statistical significance of differences in parseability across output formats, we conducted paired McNemar’s tests and report the results in Table~\ref{tab:mcnemar-results}.

\begin{table}[h!]
\centering
\renewcommand{\arraystretch}{0.8}

\resizebox{0.7\linewidth}{!}{
\begin{tabular}{lcc}
\toprule
\textbf{Comparison} & \textbf{$\chi^2$} & \textbf{p-value} \\
\midrule
JSON vs YAML & 167.607 & $\ll 0.05$ \\
JSON vs XML  & 69.351  & $\ll 0.05$ \\
YAML vs XML  & 32.411  & $\ll 0.05$ \\
\bottomrule
\end{tabular}
}
\caption{Paired McNemar’s test results comparing parseability outcomes across formats}
\label{tab:mcnemar-results}
\end{table}

All comparisons yield statistically significant results, with JSON significantly outperforming both YAML and XML ($p \ll 0.05$ in both cases). The difference between YAML and XML is also significant ($p \ll 0.05$), though comparatively smaller in effect size.

\begin{figure}[h!]
    \centering
    \includegraphics[width=0.5\textwidth]{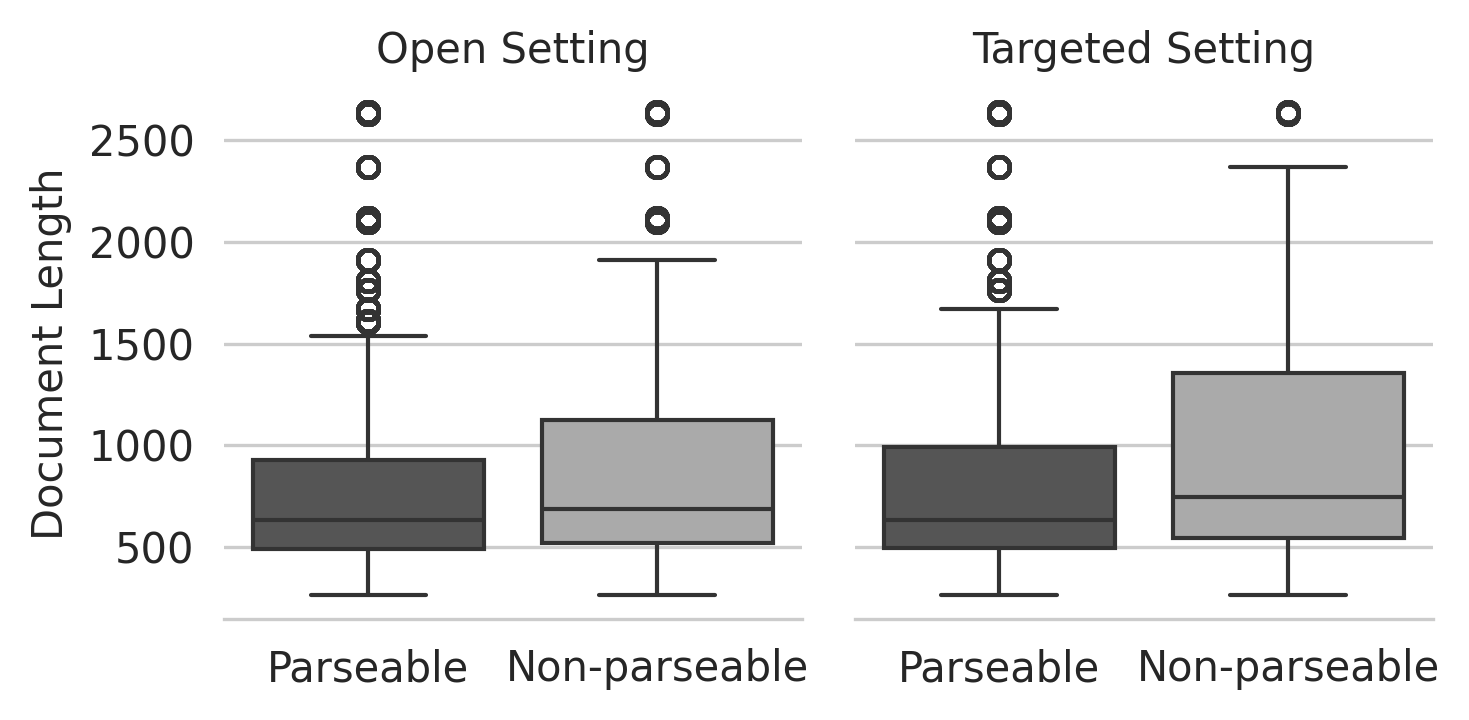}
    \caption{Boxplot showing the distribution of document lengths (in words) for parseable and non-parseable outputs.}

    \label{fig:doclen-parseable}
\end{figure}

Figure~\ref{fig:doclen-parseable} illustrates the relationship between document length (in words) and parseability, separately for the open and targeted extraction scenarios. In both scenarios, documents that failed to parse tend to be longer, with noticeably higher medians and more dispersed distributions compared to parseable documents.

To quantify the relationship between document length and parseability, we computed the point-biserial correlation. Across all documents, the correlation was weak but statistically significant (\( r = -0.081, p \ll 0.05 \)). When analyzed by scenario, the negative correlation was slightly stronger in the open setting (\( r = -0.118, p \ll 0.05 \)) compared to the targeted setting (\( r = -0.077, p \ll 0.05 \)). These results suggest that longer documents are consistently less likely to be parsed successfully, especially in open-ended generation scenarios. However, despite statistical significance, the small effect size and substantial overlap in length distributions between parseable and non-parseable documents (Figure~\ref{fig:doclen-parseable}) indicate that length alone does not strongly determine parseability. This suggests the presence of potential confounding factors such as note type, which we examine further.

\begin{figure}[h!]
    \centering
    \includegraphics[width=0.45\textwidth]{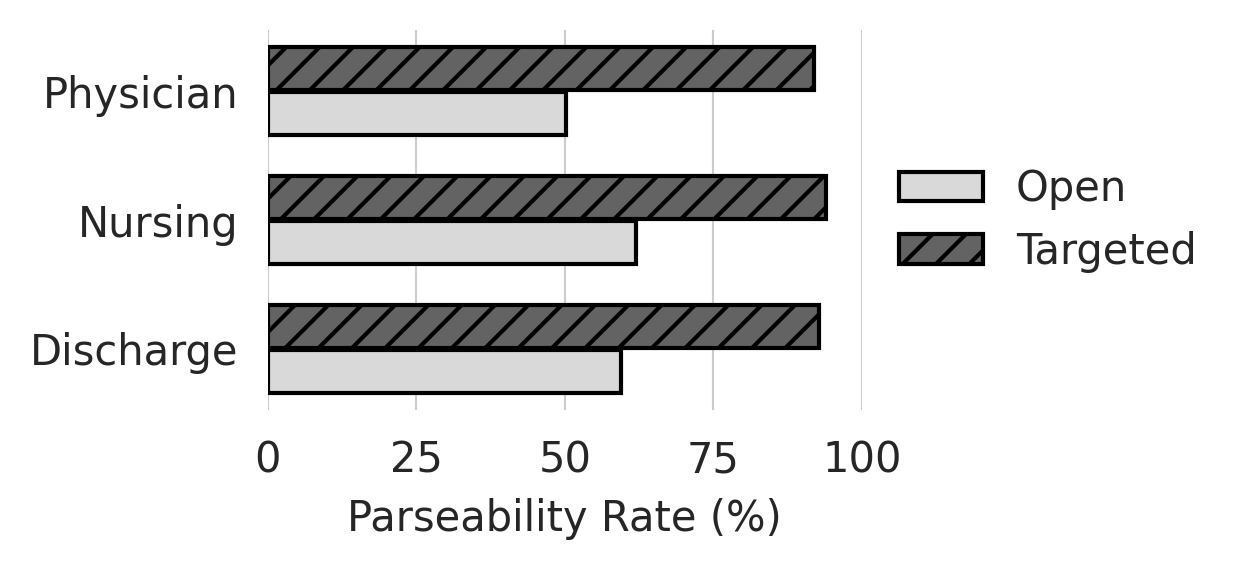}
    \caption{Parseability rates by document type for open and targeted extraction settings. Bars show the percentage of successfully parsed documents within each type.}

    \label{fig:doctype-parseable}
\end{figure}

Figure~\ref{fig:doctype-parseable} shows parseability rates across the three clinical document types, separated by extraction scenario. Targeted prompting consistently improves parseability for all types, with the most pronounced gain observed in physician notes. Nursing notes achieve the highest parseability overall, while physician notes lag behind in the open setting. These differences likely reflect variations in document complexity and length, as shown in Table~\ref{tab:note_stats}, where physician notes are among the longest on average. To assess whether document type is significantly associated with parseability, we conducted a chi-squared test of independence, yielding \( \chi^2 = 23.93 \), \( p \ll 0.05 \). This confirms that the observed differences across note types are unlikely to be due to chance, though the corresponding Cramér’s \( V = 0.05 \) indicates a small effect size.


To isolate the effects of document type and length on parseability, we fit a logistic regression with parseability as the binary outcome. Results show that discharge notes, though longer on average, are more parseable than nursing notes (\( \beta = 0.550 \), \( p < 0.05 \)), while physician notes are less parseable (\( \beta = -0.204 \), \( p < 0.05 \)). Length itself negatively impacts parseability (\( \beta = -0.0008 \), \( p < 0.05 \)). These findings suggest that document type affects parseability independently of length, likely due to semantic and structural differences. 

To understand the structural differences suggested by the regression analysis, we performed a qualitative analysis of the notes. This analysis reveals distinct structural patterns that explain these findings. Discharge notes are more consistently templated, with consistent section headers and enumerated lists that facilitate structured parsing, even in longer documents. In contrast, physician notes are rich in semantically dense content and frequently include compact representations of clinical data, such as vitals and lab panels (e.g., \textit{Ca\textsuperscript{++}: 8.3 mg/dL, Mg\textsuperscript{++}: 2.7 mg/dL, PO\textsubscript{4}: 5.0 mg/dL}), that pose specific challenges for structured formatting. These notations often combine numbers, units, and symbols in complex strings that can break parsing when not properly quoted or escaped. Nursing notes fall in between, mixing structured elements like vitals and interventions with narrative descriptions of patient events. These semantic and structural distinctions, not length alone, appear to drive parseability differences across note types.

\section{Error Analysis}

\begin{figure}[h!]
    \centering
    \includegraphics[width=0.9\linewidth]{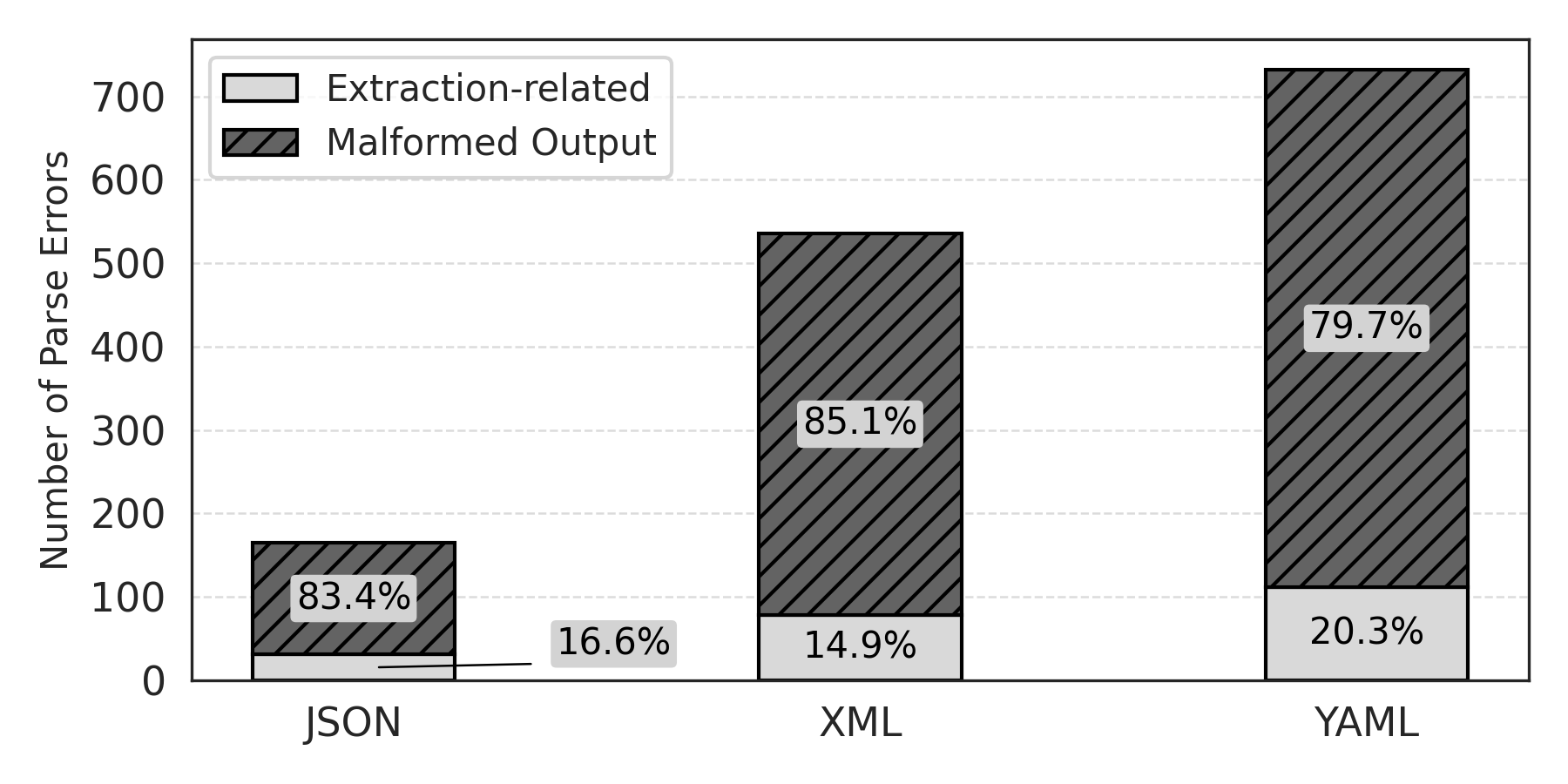}
    \caption{Breakdown of parse errors across JSON, XML, and YAML formats. Bars show the number of extraction-related and malformed output errors per format.}
    \label{fig:parse-error-breakdown}
\end{figure}

We categorize parse errors into two broad groups. First, extraction-related errors (see Figure~\ref{fig:parse-error-breakdown}, ``Extraction-related'' portion) occur when a standard regular expression fails to extract a structured object from the model output. Notably, our analysis revealed that the majority of extraction-related errors stemmed from infinite repetitions \cite{holtzman2020curious} in the generated text.

Second, malformed output errors, which arise when the output is syntactically invalid and cannot be parsed after successful extraction. Figure~\ref{fig:parse-error-breakdown} shows the distribution of these error types across formats. A more detailed breakdown is provided in Appendix~\ref{sec:appendix2}.

To quantify the association between model size and types of parse errors, we grouped failed generations by model size. Among these, Large (14B) models produced only 2.4\% extraction-related errors, compared to 21.0\% and 19.0\% for Medium (8B) and Small (3-4B) models, respectively. A Chi-squared test confirmed a statistically significant association between model size and error type ($\chi^2 = 45.52$, $p \ll 0.05$), with a Cramér's $V = 0.18$ indicating a small to moderate effect size. These findings suggest that extraction errors are more typical in smaller models, though they are not exclusive to them.

We also examined whether the type of parse error varied with prompt type. Open prompts resulted in extraction-related errors only 2.4\% of the time, while targeted prompts produced extraction errors in 45.5\% of failures. A Chi-squared test revealed a statistically significant association between prompt type and error type ($\chi^2 = 420.62$, $p \ll 0.05$), and Cramér's $V = 0.54$ indicated a large effect size. This suggests that extraction errors are a dominant failure mode under targeted prompting conditions.

\section*{Conclusion}
We conducted a systematic evaluation of the structural robustness of SLM-generated outputs for open attribute-value extraction from clinical notes. Across three common formats, JSON significantly outperformed YAML and XML in parseability. Parseability improved with model size and prompt specificity, and targeted prompting yielded especially large gains for YAML. However, performance declined on longer documents, and physician notes were particularly error-prone. Error analysis revealed two dominant failure modes: infinite repetition and syntactic malformations, particularly missing quotation marks around numerals embedded in non-numeric fields (e.g., blood pressure values like ``128/68''), unescaped special characters, and malformed list structures. These issues were most frequent in smaller models and underscore the need for decoding strategies that promote format-conformant output.

Our findings underscore the importance of aligning prompt and format design with generation strategies that ensure structural reliability, particularly in resource-constrained or privacy-sensitive clinical NLP settings. Future work should explore automatic post-processing techniques to detect and correct structural errors, extend parsers to better handle common irregularities in LLM-generated outputs, conduct more extensive evaluations on diverse clinical corpora, and support joint analysis of syntactic and semantic validity to better assess the clinical utility of structured outputs.

\section*{Limitations}

While our study offers detailed insights into the structural robustness of SLM outputs, it has several limitations. First, the evaluation is based on the EHRCon dataset, which, although diverse in note types, contains only 105 documents and may not capture the full variability of clinical narratives. Second, all experiments were conducted using a single decoding configuration (greedy decoding without sampling), which may not generalize to alternative generation settings. Third, we evaluated a limited set of open-weight models. Future work should include domain-specific clinical language models and additional parameter sizes to capture broader trends. Finally, our analysis focused exclusively on syntactic parseability, without assessing the semantic accuracy or clinical correctness of the extracted information, which is an important direction for future research.

\section*{Ethics Statement}

This study uses the EHRCon dataset, which is derived from the publicly available and de-identified MIMIC-III database. As no personally identifiable information is included in the data, and no new data collection was conducted, the study does not require approval from an institutional ethics board. We do not publish any content that could potentially identify individuals. To promote transparency and reproducibility, we rely exclusively on open-source models and datasets, and provide detailed descriptions of our experimental setup and evaluation methodology.

\bibliography{custom}

\newpage
\appendix

\raggedbottom

\section{Additional Details on Experimental Setup}
\label{sec:appendix1}

\subsection*{Software Versions}

Experiments were conducted using Python 3.10.12 (main, Nov 20 2023, 15:14:05) compiled with GCC 11.4.0. Table~\ref{tab:software-versions} lists the versions of key libraries used in our experiments.

\begin{table}[h]
\centering
\resizebox{0.6\linewidth}{!}{
\begin{tabular}{@{}ll@{}}
\toprule
\textbf{Library} & \textbf{Version} \\
\midrule
transformers & 4.51.3 \\
PyYAML & 6.0.1 \\
statsmodels & 0.14.2 \\
scipy & 1.13.1 \\
numpy & 1.26.4 \\
json & Standard Library (Python 3.10) \\
xml & Standard Library (Python 3.10) \\
\bottomrule
\end{tabular}
}
\caption{Versions of software and libraries used in the experiments.}
\label{tab:software-versions}
\end{table}

\subsection*{Model Configuration}

All models were queried using the HuggingFace pipeline interface with parameters listed in Table~\ref{tab:model-config}. Generation was deterministic and capped at 8192 tokens. For consistency across models, the ``thinking'' mode was disabled for Qwen models.

\begin{table}[h]
\centering
\resizebox{0.4\linewidth}{!}{
\begin{tabular}{@{}ll@{}}
\toprule
\textbf{Parameter} & \textbf{Value} \\
\midrule
\texttt{max\_new\_tokens} & 8192 \\
\texttt{do\_sample} & False \\
\texttt{top\_p} & None \\
\texttt{temperature} & None \\
\bottomrule
\end{tabular}
}
\caption{Model generation parameters used in all decoding runs.}
\label{tab:model-config}
\end{table}

\subsection*{Regular Expressions}

If initial parsing failed, we attempted to extract structured content from fenced code blocks using regular expressions. Table~\ref{tab:regex-formats} summarizes the patterns used for each format.

\begin{table}[h]
\centering
\resizebox{\linewidth}{!}{
\begin{tabular}{@{}llp{0.55\linewidth}@{}}
\toprule
\textbf{Format} & \textbf{Regex Pattern} & \textbf{Description} \\
\midrule
JSON & \texttt{\detokenize{```(?:json)?\s*\n(.*?)```}} & Matches a fenced code block optionally labeled as \texttt{json}. Extracts everything between the triple backticks. \\
YAML & \texttt{\detokenize{```(?:yaml|yml)?\s*\n(.*?)```}} & Matches a fenced code block optionally labeled as \texttt{yaml} or \texttt{yml}. Captures the inner content. \\
XML  & \texttt{\detokenize{```(?:xml)?\s*\n(.*?)```}} & Matches a fenced code block optionally labeled as \texttt{xml}. Content inside is captured for parsing. \\
\bottomrule
\end{tabular}
}
\caption{Regular expressions used to extract structured content from fenced code blocks.}
\label{tab:regex-formats}
\end{table}

\subsection*{Prompts}

For open-ended attribute-value extraction, we used format-specific prompts that instructed the model to generate structured data in either JSON, YAML, or XML. Each prompt asked the model to produce a valid, well-structured output using the appropriate syntax and meaningful field names. Additionally, models were explicitly instructed to use proper serialization fences to support regex-based extraction.

The general prompt template is shown below, where \texttt{<FORMAT>} is replaced with the target format (\texttt{JSON}, \texttt{YAML}, or \texttt{XML}):

\begin{tcolorbox}[colback=gray!5!white, colframe=gray!75!black, title=Open Extraction Prompt]
Given the following document: \textbackslash n \textit{<document text>}. Extract all data in \texttt{<FORMAT>} format.  
Make sure that the \texttt{<FORMAT>} document is valid, provide reasonably detailed names for fields.  
  
Make a proper fence for \texttt{<FORMAT>} so that it can be extracted from the response with a regular expression.
\end{tcolorbox}

For targeted extraction scenario, we used prompts that explicitly instructed the model to extract specific categories: demographics, medications, or symptoms, in a specified structured format. Prompts were adjusted dynamically based on both the target concept and the desired output format (JSON, YAML, or XML). If no relevant information was found, the model was instructed to return an empty object.

The generalized prompt template is shown below, where \texttt{<CONCEPT>} refers to the target category (e.g., ``patient demographics'' or ``medications'') and \texttt{<FORMAT>} specifies the output format.

\begin{tcolorbox}[colback=gray!5!white, colframe=gray!75!black, title=Targeted Extraction Prompt]
Given the following document: \textbackslash n \textit{<document text>}.
Extract all mentioned \texttt{<CONCEPT>} from the text below in valid \texttt{<FORMAT>} format.  
If no \texttt{<CONCEPT>} are found, return an empty \texttt{<FORMAT>} object.
Make sure that the \texttt{<FORMAT>} document is valid, provide reasonably detailed names for fields.  

Make a proper fence for \texttt{<FORMAT>} so that it can be extracted from the response with a regular expression.

\end{tcolorbox}

\section{Additional Details on Error Analysis}
\label{sec:appendix2}

\subsection{Extraction-Related Errors}

Extraction-related errors arise when neither direct parsing nor regular expression matching succeeds in recovering a structured object from the model output. Initially, we attempt to parse the output as-is, assuming the model produces a complete structured object without serialization fences; if that fails, we apply format-specific regular expressions to extract fenced content (Appendix~\ref{sec:appendix1}). These errors predominantly stem from infinite repetitions in the generated text. Table~\ref{tab:extraction_failures} summarizes the extraction-related failures across all formats. Notably, \texttt{Phi-4} was the only model that consistently avoided these failures.

\begin{table}[h!]
\centering
\resizebox{1\linewidth}{!}{
\begin{tabular}{lccc}
\toprule
\textbf{Format} & \textbf{Total Cases} & \textbf{Infinite Repetitions} & \shortstack{\textbf{Broken Fence} \\ \textbf{(Non-repetitive)}} \\
\midrule
JSON  & 31  & 31  & 0 \\
XML   & 78  & 78  & 0 \\
YAML  & 112 & 109 & 3 \\
\bottomrule
\end{tabular}
}
\caption{Summary of extraction-related failures due to regular expression mismatches.}
\label{tab:extraction_failures}
\end{table}

The repetition block length varied, ranging from short fragments such as:

\begin{tcolorbox}[colback=gray!10, colframe=gray!60, boxrule=0.5pt, arc=2pt, left=2pt, right=2pt, top=2pt, bottom=2pt]
\footnotesize
"Hepatic dysfunction",\\
"Hepatic dysfunction",\\
"Hepatic dysfunction",\\
"Hepatic dysfunction",\\
"Hepatic dysfunction"
\end{tcolorbox}

to much longer blocks like:

\begin{tcolorbox}[colback=gray!10, colframe=gray!60, boxrule=0.5pt, arc=2pt, left=2pt, right=2pt, top=2pt, bottom=2pt]
\footnotesize
"shortness of breath or respiratory distress (not explicitly stated but implied by SpO2: 100\%)",\\
"chest pain or discomfort (not explicitly stated but implied by clear lungs on CXR)",\\
"fever or chills (not explicitly stated but implied by WBC: 12.4 and 13.8)",\\
"abdominal pain or discomfort (epigastric region)",\\
"nausea or vomiting (not explicitly stated but implied by NPO status)",\\
"abdominal distension (nondistended)",\\
"abdominal tenderness (TTP in all quadrants)",\\
"abdominal guarding (voluntary guarding)",\\
"abdominal masses or organomegaly (not explicitly stated but implied by TTP in all quadrants)",\\
"shortness of breath or respiratory distress (not explicitly stated but implied by SpO2: 100\%)",\\
"chest pain or discomfort
\end{tcolorbox}

\subsection{Malformed Output Errors}

Malformed output errors occur when the internal content of a model's generation is structurally invalid, resulting in failed parsing despite the successful extraction of the object. 

Because these issues are tightly coupled to the specific requirements of each format, we analyze them separately for JSON, XML, and YAML.

Table~\ref{tab:json-errors} summarizes the most common sources of malformed JSON, including unquoted values, missing delimiters, improperly structured lists, and misnested objects. Many of these errors stem from the model emitting raw numerical data, units, or complex expressions without enclosing them in quotes.

Table~\ref{tab:xml-errors} highlights XML-specific issues such as invalid tag names, unescaped reserved characters (e.g., \texttt{\&}, \texttt{<}), and improper tag nesting. Additional problems arise when tags encode entire phrases or when outputs terminate prematurely, leaving the structure incomplete.

Table~\ref{tab:yaml-errors} details YAML parsing failures, which are frequently caused by incorrect use of aliases, inconsistent indentation, missing colons, or unescaped colons within long strings. YAML is particularly sensitive to formatting errors, making  minor deviations from proper structure likely to result in failure.

\begin{table*}[h]
\centering
\small
\begin{tabular}{p{3cm}p{6cm}p{6cm}}
\toprule
\textbf{Category} & \textbf{Description} & \textbf{Example} \\
\midrule
Unquoted numeric values & Common vitals (e.g., \texttt{128/68}, \texttt{96\%}) were emitted without quotes, causing syntax errors. & \verb|"blood_pressure": 128/68,| \\
Unquoted units or ranges & Values with units (\texttt{300mg}, \texttt{20-60cc/hr}) appeared as raw text. & \verb|"dose": 300mg,| \\
Improper list or array formatting & Lists with non-JSON-safe elements (e.g., slashed values) were incorrectly serialized. & \verb|"BP": [121/63, 75],| \\
String concatenation or unescaped expressions & Attempted concatenation or strings with internal quotes broke JSON structure. & \verb|"Range": "10 - 20" + " insp/min",| \\
Missing delimiters & Adjacent fields were emitted without commas. & \verb|"hematocrit": 37.3 "platelets": 126 K,| \\
Standalone strings & Free text like \texttt{"Levofloxacin"} appeared without a key, resembling list items. & 
\begin{minipage}[t]{4cm}
\begin{verbatim}
"medications": {
  "Levofloxacin"
}
\end{verbatim}
\end{minipage} \\
Multiple top-level objects & More than one top-level JSON object or extraneous content after the main object. & 
\begin{minipage}[t]{4cm}
\begin{verbatim}
{
  "History": "..."
}
{
  "PMH": {...}
}
\end{verbatim}
\end{minipage} \\
Unescaped control characters & Strings included invalid characters or unmatched quotes. & 
\begin{minipage}[t]{4cm}
\begin{verbatim}
"date": "s/p lobectomy '[**33**]'
\end{verbatim}
\end{minipage} \\
\bottomrule
\end{tabular}
\caption{Summary of prevalent JSON formatting errors in model outputs.}
\label{tab:json-errors}
\end{table*}

\begin{table*}[h]
\centering
\small
\begin{tabular}{p{2cm}p{3cm}p{10cm}}
\toprule
\textbf{Category} & \textbf{Description} & \textbf{Example} \\
\midrule
Invalid tag names & Tags contain digits, punctuation, or special characters, violating XML naming rules. & \verb|<123_BP>120/80</123_BP>| \\
Unescaped characters & Raw XML-reserved characters (\texttt{<}, \texttt{>}, \texttt{\&}) appear unescaped in text content. & \verb|<symptom>nausea & vomiting</symptom>| \\
Mismatched or misnested tags & Opening and closing tags are misaligned or improperly nested. & \verb|<heart><rate>88</heart></rate>| \\

Improper structural nesting & Structural templates are reused in invalid contexts or nested inconsistently. & \verb|<24_hour_events><note>...</24_hour_events></note>| \\
Free-text as tag name & Sentence-length strings or clinical statements are incorrectly placed as tag names. & \verb|<Patient is alert and oriented>yes</Patient is alert and oriented>| \\

\bottomrule
\end{tabular}
\caption{Summary of prevalent XML formatting errors in model outputs.}
\label{tab:xml-errors}
\end{table*}

\begin{table*}[h]
\centering
\small
\begin{tabular}{p{2.5cm}p{3cm}p{9.5cm}}
\toprule
\textbf{Category} & \textbf{Description} & \textbf{Example} \\
\midrule
Alias misinterpretation & Placeholders in \texttt{[**...**]} format are misinterpreted as YAML aliases, which require alphanumeric characters. & \verb|attending_md: [**Doctor Last Name**] [**Doctor First Name**] C.| \\
Invalid nested mappings & Multiple colons in a single line without proper quoting create ambiguous mappings. & \verb|- Cardiovascular: (S1: Normal), (S2: Normal)| \\
Improper scalar values & Misuse of block scalars (e.g., \texttt{>}) or unescaped strings leads to format violations. & \verb|- SpO2: >95\%| \\
Unclosed or broken blocks & Incomplete sequences or mappings with missing indentation or block terminators. & \verb|- Fentanyl: "2192-9-17" 08:10 AM| \\
Malformed collections & Lists with poor indentation or unexpected formatting cannot be resolved by the parser. & \verb|- "not feeling well" (1 day prior to admission)| \\
Improper question mark usage & Use of \texttt{?} outside mapping syntax breaks YAML interpretation. & \verb|?look into the suprapubic area.| \\
Unescaped strings with colons & Long unquoted strings containing multiple colons (e.g., copied EHR text) are misparsed. & \verb|title: Chief Complaint: respiratory failure, PEA arrest| \\
\bottomrule
\end{tabular}
\caption{Summary of prevalent YAML formatting errors in model-generated outputs.}
\label{tab:yaml-errors}
\end{table*}

\end{document}